\documentclass[sn-mathphys,Numbered]{sn-jnl}

\usepackage{amsmath,amssymb,amsfonts}%
\usepackage{amsthm}%
\usepackage{mathrsfs}%
\usepackage[title]{appendix}%
\usepackage{textcomp}%
\usepackage{manyfoot}%
\usepackage{algorithm}%
\usepackage{algorithmicx}%
\usepackage{algpseudocode}%
\usepackage{listings}%

\usepackage{graphicx}
\usepackage{amsmath}
\usepackage{amssymb}
\usepackage{booktabs}
\usepackage{multirow}
\usepackage{tabularx}
\usepackage{algorithm}
\usepackage{pifont}
\usepackage{balance}
\usepackage{makecell}
\usepackage{epsfig}
\usepackage{amsmath}
\usepackage{amssymb}
\usepackage{comment}
\usepackage{multirow}
\usepackage{tabularx}
\usepackage{balance}
\usepackage{balance}
\usepackage{algorithm}
\usepackage{xcolor}
\usepackage{color, colortbl}
\usepackage{pdfpages}
\usepackage{caption}
\usepackage{subcaption}
\usepackage{enumitem}
\usepackage{diagbox}

\newcommand{\highlight}[1]{\textcolor{black}{#1}}

\definecolor{lightgray}{gray}{0.9}

\def\onedot{.}
\def\eg{\emph{e.g}\onedot} 
\def\ie{\emph{i.e}\onedot}

\def\etal{\emph{et al}\onedot}

\theoremstyle{thmstyleone}%
\theoremstyle{thmstyletwo}%

\theoremstyle{thmstylethree}%

\begin{document}

\title[Article Title]{GUNNEL: Guided Mixup Augmentation and Multi-Model Fusion for Aquatic Animal Segmentation}


\author[1,2]{\fnm{Minh-Quan} \sur{Le}}\email{lmquan@selab.hcmus.edu.vn}\equalcont{These authors contributed equally to this work.} 
\author*[1,2]{\fnm{Trung-Nghia} \sur{Le}}\email{ltnghia@fit.hcmus.edu.vn}\equalcont{These authors contributed equally to this work.} 
\author[3]{\fnm{Tam V.} \sur{Nguyen}}\email{tamnguyen@udayton.edu}
\author[4]{\fnm{Isao} \sur{Echizen}}\email{iechizen@nii.ac.jp}
\author[1,2]{\fnm{Minh-Triet} \sur{Tran}}\email{tmtriet@fit.hcmus.edu.vn}

\affil[1]{\orgname{Vietnam National University}, \orgaddress{\postcode{700000}, \state{Ho Chi Minh}, \country{Vietnam}}}
\affil[2]{\orgname{University of Science}, \orgaddress{\street{227 Nguyen Van Cu}, \postcode{700000}, \state{Ho Chi Minh}, \country{Vietnam}}}
\affil[3]{\orgname{University of Dayton}, \orgaddress{\street{300 College Park}, \city{Dayton}, \postcode{45469}, \state{Ohio}, \country{US}}}
\affil[4]{\orgname{National Institute of Informatics}, \orgaddress{\street{2-1-2 Hitotsubashi}, \city{Chiyoda-ku}, \postcode{1018043}, \state{Tokyo}, \country{Japan}}}


\abstract{Recent years have witnessed great advances in object segmentation research. In addition to generic objects, aquatic animals have attracted research attention. Deep learning-based methods are widely used for aquatic animal segmentation and have achieved promising performance. However, there is a lack of challenging datasets for benchmarking. In this work, we build a new dataset dubbed ``Aquatic Animal Species." We also devise a novel \textbf{GU}ided mixup augme\textbf{N}tatio\textbf{N} and multi-mod\textbf{E}l fusion for aquatic anima\textbf{L} segmentation (GUNNEL) that leverages the advantages of multiple segmentation models to segment aquatic animals effectively and improves the training performance by synthesizing hard samples. Extensive experiments demonstrated the superiority of our proposed framework over existing state-of-the-art instance segmentation methods. 
}



\maketitle

\section{Introduction}


\label{sec:intro}
Computer vision is an interdisciplinary research field that helps computers gain understanding from visual data sources. The rapid development of deep learning techniques has enabled computer vision applications to become a part of daily life, such as autonomous driving~\cite{ltnghia-IV2020}, medical treatment~\cite{Sainz-MICCAI2020}, entertainment~\cite{vmduc-WACV2018}, shopping~\cite{Nguyen-PACIS2012}, surveillance~\cite{tmtriet-AICity2020, Nguyen-FG2021}, and wildlife preserve~\cite{ltnghia-CVIU2019}. These applications require detecting and segmenting various types of objects appearing in images and videos~\cite{Liu-IJCV2019, Minaee-TPAMI2021}.


Animal-related research has recently drawn the attention of the computer vision community. As a result, the number of studies involving animal research using computer vision and deep learning techniques has been increasing, such as measuring animal behaviors~\cite{Mathis-CPN2020} and analyzing animal trajectories~\cite{Maekawa-NC2020}. In addition, various animal-related datasets have been constructed for specific research purposes~\cite{parkhi-CVPR2012, Wah-CUB2011, Wu-CVPR2019}. However, there is a lack of research on aquatic animals, especially on automatically localizing and recognizing aquatic animals. We attribute this partially to the lack of appropriate datasets for training, testing, and benchmarking. Thus, constructing datasets of aquatic animals has become essential to develop methods for detecting and segmenting aquatic animal instances. 

We created a dataset to encourage more research on aquatic animal segmentation, namely ``Aquatic Animal Species (AAS)." Our dataset consists of $4,239$ images with $5,041$ instances of 46 aquatic animal species. We also devised a simple yet effective \textbf{GU}ided mixup augme\textbf{N}tatio\textbf{N} and multi-mod\textbf{E}l fusion for aquatic anima\textbf{L} segmentation (GUNNEL). In our proposed multi-model fusion method, the results of multiple instance segmentation models are fused under the guidance of a mask-agnostic controller. The single models exploit different contexts to create different segmentation masks for the aquatic animals in the image. The model controller aims to produce pseudo-bounding box ground truths for fusing multiple instance segmentation models. Moreover, to overcome issues of limited training data, we developed a guided mixup augmentation method based on a confusion matrix to improve training performance. We further established the first comprehensive benchmark for aquatic animal segmentation. Extensive experiments on the newly constructed dataset demonstrated the superiority of our proposed method over state-of-the-art instance segmentation methods. 

Our contributions are five-fold:
\begin{itemize}
    \item We present a comprehensive study on aquatic animal segmentation, which is more complicated than general object segmentation. This work introduces a new benchmark for the task of aquatic animal segmentation.

    \item We propose a new multi-model fusion strategy for aquatic animal segmentation. Our fusion scheme leverages different instance segmentation models by using multi-model mask fusion under the guidance of a mask-independent controller. To the best of our knowledge, our proposed fusion strategy is the first ensemble method for the instance segmentation task.
    
    \item We develop a new guided mixup augmentation method for improving segmentation performance. Our augmentation strategy boosts the performance of deep learning models from a discriminative ability perspective and maintains the global context, especially for camouflaged aquatic animals.
    
    \item We construct a new dataset for aquatic animal segmentation. Our newly constructed AAS dataset contains $4,239$ images of 46 aquatic animal species, each image having 1.2 instances on average. All images have instance-wise ground truths of the bounding box and segmentation mask.

    \item We introduce a comprehensive AAS benchmark to support advancements in this research field.
    
\end{itemize}

\highlight{The remainder of this paper is organized as follows. Section~\ref{sec:related_work} summarizes the related work. Next, Section~\ref{sec:proposed_dataset} presents our newly constructed AAS dataset. Our proposed GUNNEL framework for aquatic animal segmentation is introduced in Section~\ref{sec:proposed_method}. Section~\ref{sec:experiments} reports the evaluation and in-depth analysis of our proposed method. Finally, Section~\ref{sec:conclusion} draws the conclusion and paves the way for future work.}

\section{Related Work}
\label{sec:related_work}


\subsection{Instance Segmentation} 

Instance segmentation methods not only localize objects but also predict per-pixel segmentation masks of objects with their corresponding semantic labels. Mask RCNN~\cite{Kaiming-ICCV2017}, which is known as a pioneer end-to-end deep neural network for instance segmentation, and its variances (\eg, Cascade Mask RCNN~\cite{Zhaowei-CVPR2018}, Mask Scoring RCNN~\cite{Huang-CVPR2019}) perform detect-then-segment. After bounding box object detection, these models involve object mask segmentation by adding segmentation branches on top of Faster RCNN~\cite{Ren-NIPS2015}. Cascade Mask RCNN~\cite{Zhaowei-CVPR2018, Hafiz-IJMIR2020} and Mask Scoring RCNN~\cite{Huang-CVPR2019} were developed to solve issues of the intersection over union (IoU) threshold by training a multi-stage architecture and evaluating the quality of mask prediction.

Recent efforts have explored instance segmentation using ideas from other computer vision and computer graphics branches~\cite{Jiaqi-ICCV2019, Xizhou-CVPR2019, Cao-CVPRW2019, Kirillov-CVPR2020}. Wang~\etal~\cite{Jiaqi-ICCV2019} designed a universal, lightweight, and highly effective feature upsampling operator. Zhu~\etal~\cite{Xizhou-CVPR2019} improved the ability to concentrate on pertinent image regions through comprehensive integration of deformable convolutions. Cao~\etal~\cite{Cao-CVPRW2019} proposed a global context network by applying query-independent formulation into network layers. Kirillov~\etal ~\cite{Kirillov-CVPR2020} developed a rendering-like segmentation module, which performs point-based segmentation at adaptively selected locations based on an iterative subdivision algorithm. Rossi~\etal~\cite{Rossi-ICPR2021} applied non-local building blocks and attention mechanisms to overcome the limitation of RoI extractors. Different normalization layers, such as Group Normalization~\cite{Wu-ECCV2018} and Weight Standardization~\cite{Siyuan-2019}, have been proposed for reducing gradient vanishing to improve the training performance.

The fusion of single segmentation models, dubbed ensemble modeling, is important in improving performance. Indeed, under circumstances when real-time inference is not required, ensemble modeling can improve robustness and effectiveness. Several object detection techniques have been introduced that combine the predictions of multiple models and obtain a final set of detection results. They include non-maximum suppression (NMS)~\cite{Dalal-CVPR2005, Girshick-CVPR2014}, soft-NMS~\cite{Bodla-ICCV2017}, and weighted boxes fusion~\cite{SOLOVYEV-IVC2021}. To the best of our knowledge, a fusion technique for the instance segmentation task has not been developed. Here, we present a novel fusion algorithm leveraging a simple detector that performs as a model controller to ensemble predictions of instance segmentation models.

\subsection{Data Augmentation} 
\label{sec:related_work_augmentation}

Data augmentation consists of a collection of transformation techniques that inflate the size and quality of training datasets. These techniques improve the generalization of deep learning models~\cite{Shorten-JBD2019}. Through data augmentation, deep learning models can extract additional information from a dataset by data warping or oversampling. Data warping augmentation methods, which include geometric and color transformations, are always used in training deep learning models (\eg scaling, flipping, rotation, random crop~\cite{Takahashi-TCSVT2020}, color jittering~\cite{Szegedy-CVPR2015}).

On the other hand, oversampling augmentation methods~\cite{zhang-ICLR2018, Goodfellow-NIPS2014} synthesize objects and then integrate them into the original training dataset. Zhang~\etal~\cite{zhang-ICLR2018} proposed mixup augmentation by creating convex combinations of pairs of samples and their labels to train neural networks. Yun~\etal~\cite{Yun-ICCV2019} introduced CutMix mechanism to overcome the model confusion of localization of the mixup method. Bochkovskiy~\etal~\cite{Bochkovskiy-arxiv2020} developed Mosaic technique by mixing different contextual images to make object detectors conscious of objects outside their usual contexts and smaller scale as well.

In addition, recent object-aware augmentation methods~\cite{Fang-ICCV2019, Ghiasi-CVPR2021} aim to copy and paste objects from an image into another image. Dvornik~\etal~\cite{Dvornik-ECCV2018} leveraged segmentation annotations and modeled visual context surrounding objects to place copied objects. Fang~\etal~\cite{Fang-ICCV2019} explored a location probability map and found places having similar local appearances to place randomly jittered objects. Dwibedi~\etal~\cite{Dwibedi-ICCV2017} proposed to cut objects and paste them into diverse background scenes. Ghiasi~\etal~\cite{Ghiasi-CVPR2021} constructed a simple copy-paste mechanism by randomly selecting a subset of objects from images and pasting them onto others.

\begin{figure}[t!]
    \centering
    \includegraphics[width=\linewidth]{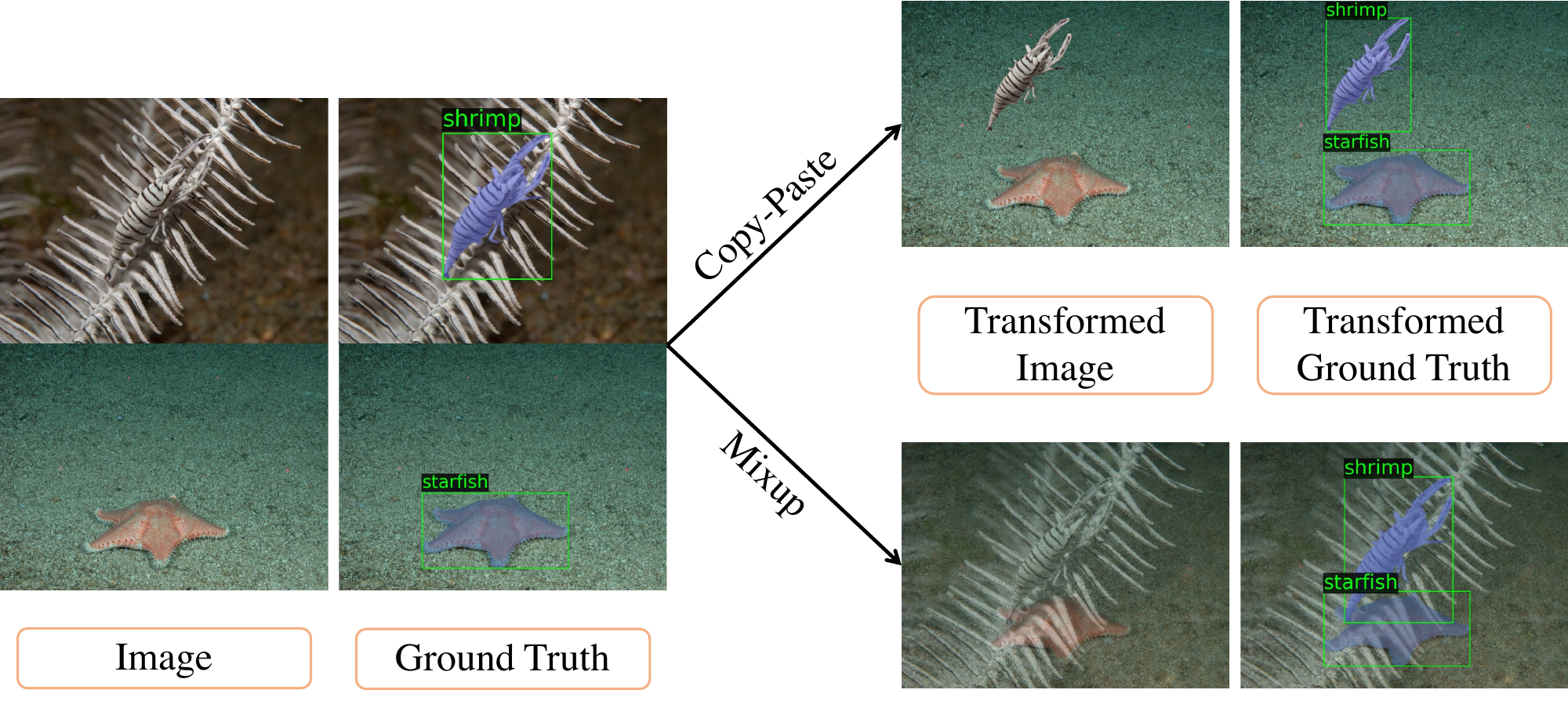}
    \caption{Comparison between copy-paste~\cite{Ghiasi-CVPR2021} and mixup~\cite{zhang-ICLR2018} data augmentations. The copy-paste method is not helpful because camouflaged aquatic animals become visible, while the mixup method transforms entire images, outputting unrealistic results (best viewed online in color with zoom). Our proposed guided mixup augmentation utilizes both approaches to achieve better performance.}
    \label{fig:copypaste_mixup}
\end{figure}

However, existing data augmentation methods do not work well on images of aquatic animals. Aquatic animal images in our AAS dataset contain both camouflaged and non-camouflaged animals. This makes the localization and recognition of an aquatic animal much more complex than animals in other datasets. In addition, identifying a camouflaged aquatic animal strongly depends on the background. This means that simply placing a cloned camouflaged aquatic animal on a random background does not work. Indeed, within different contexts of scenes, the camouflaged aquatic animal becomes far easier to be recognized~\cite{ltnghia-CVIU2019} (see Figure~\ref{fig:copypaste_mixup}).

Our proposed guided mixup augmentation method overcomes this limitation better than existing augmentation methods for three reasons. (1) Rather than randomly choosing and blending two images, a confusion matrix is leveraged to guide the model to concentrate on misclassified categories. (2) The augmentation strategy helps instance segmentation models improve their discriminative ability by choosing instances belonging to strongly confusing image categories. (3) Guided mixup augmentation helps preserve the global context for both camouflaged and non-camouflaged aquatic animals.

\subsection{Marine Organism Detection}
Wang~\etal~\cite{Wang-Neurocomputing2023} presented a comprehensive survey on deep learning-based visual detection of marine organisms. The study highlights the various deep learning models employed, such as Convolutional Neural Networks (CNNs), Recurrent Neural Networks (RNNs), and Generative Adversarial Networks (GANs). It emphasizes the challenges associated with underwater imaging, such as poor visibility, varying lighting conditions, and the complex background environment. The survey also discusses the importance of large annotated datasets and the role of data augmentation techniques in improving model performance. 

Yu~\etal\cite{Yu-AI2023} proposed a novel deep learning architecture called the Multiple Attentional Path Aggregation Network (MAPAN) for marine object detection. This model introduces a multi-path attention mechanism aggregating features from different layers to enhance the detection performance. The study demonstrates that MAPAN significantly outperforms traditional single-path models to detect marine objects, including fish, coral, and other underwater organisms. Cheng~\etal~\cite{Cheng-TCSVT2023} introduced the Bidirectional Collaborative Mentoring Network (BCMN), designed to enhance the detection of marine organisms through a novel mentoring mechanism. This approach involves two networks that collaboratively learn from each other, improving their detection capabilities. The bidirectional mentoring process allows the networks to refine their predictions iteratively, resulting in higher accuracy and reliability.



\section{Aquatic Animal Species Dataset}
\label{sec:proposed_dataset}

\begin{table}[t!]
\caption{Statistics for animal image datasets used in our experiments.}
\centering
\label{table:datasets}
\begin{tabular}{lcccc}
\hline
\noalign{\smallskip}
\textbf{Dataset} & \textbf{\#Cat.} & \textbf{\#Train Img.} & 
\textbf{\#Test Img.} & \textbf{\#Img. per Cat.} \\ 
\noalign{\smallskip}
\hline
\noalign{\smallskip}
MAS3K\cite{Li-Mas3K2020} & 37 & 1,488 & 1,007 & 67.4 \\ 
COD10K (Subset)~\cite{Fan-CVPR2020} & 21 & 758 & 293 & 50.0 \\ 
CAMO++ (Subset)\cite{ltnghia-TIP2021} & 32 & 366 & 166 & 16.6 \\ 
\rowcolor{lightgray} 
Collected AAS & 46 & 2,582 & 1,657 & 92.1 \\ \hline
\end{tabular}
\end{table}

\begin{figure}[t!]
    \centering
    \includegraphics[width=\linewidth]{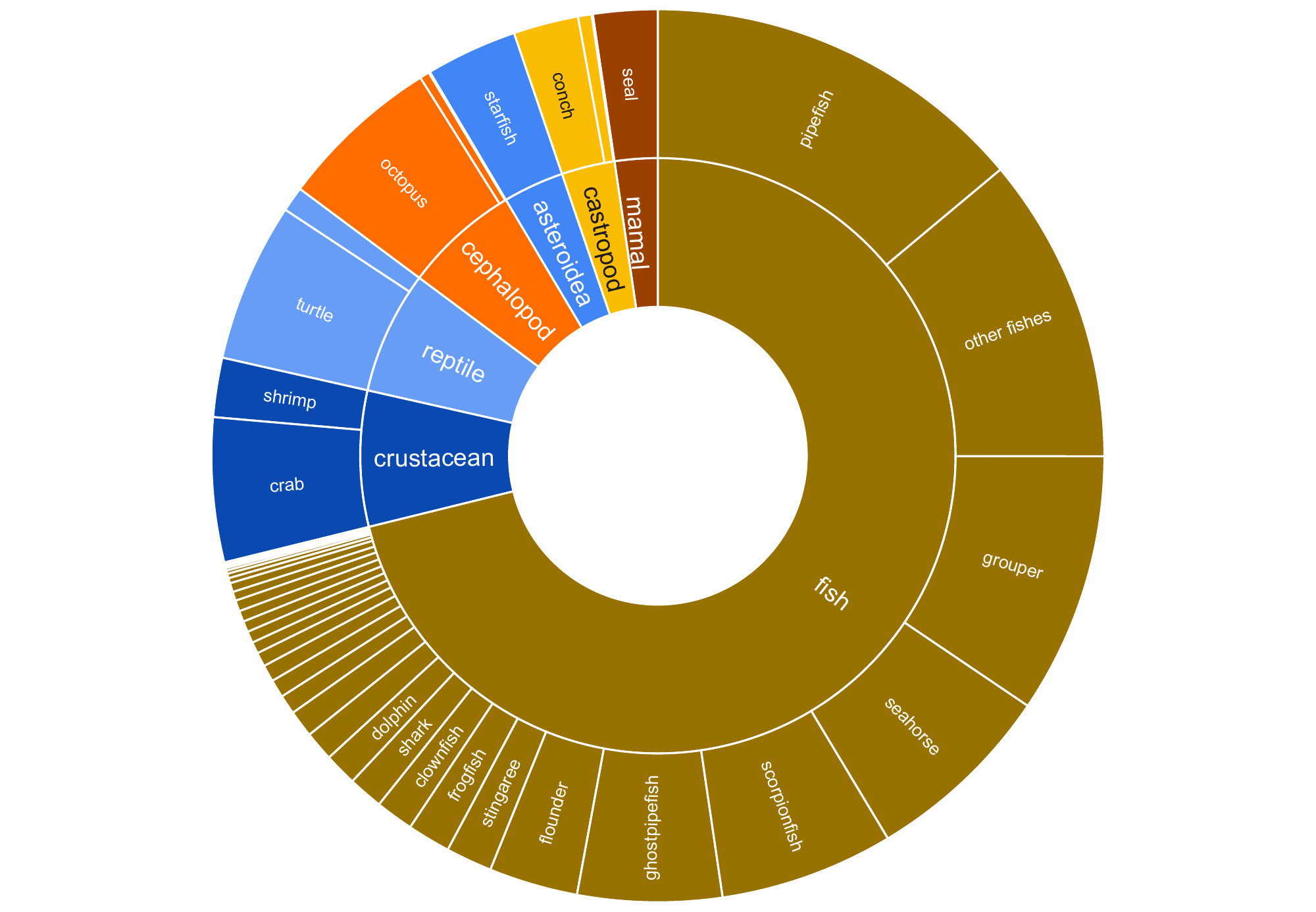}
    \caption{Aquatic animal categories in our AAS dataset (best viewed online in color with zoom). For clear presentation, we do not visualize categories with a small number of images.}
    \label{fig:animal_category}
\end{figure}

\begin{figure}[t!]
    \centering
    \includegraphics[width=\linewidth,page=1]{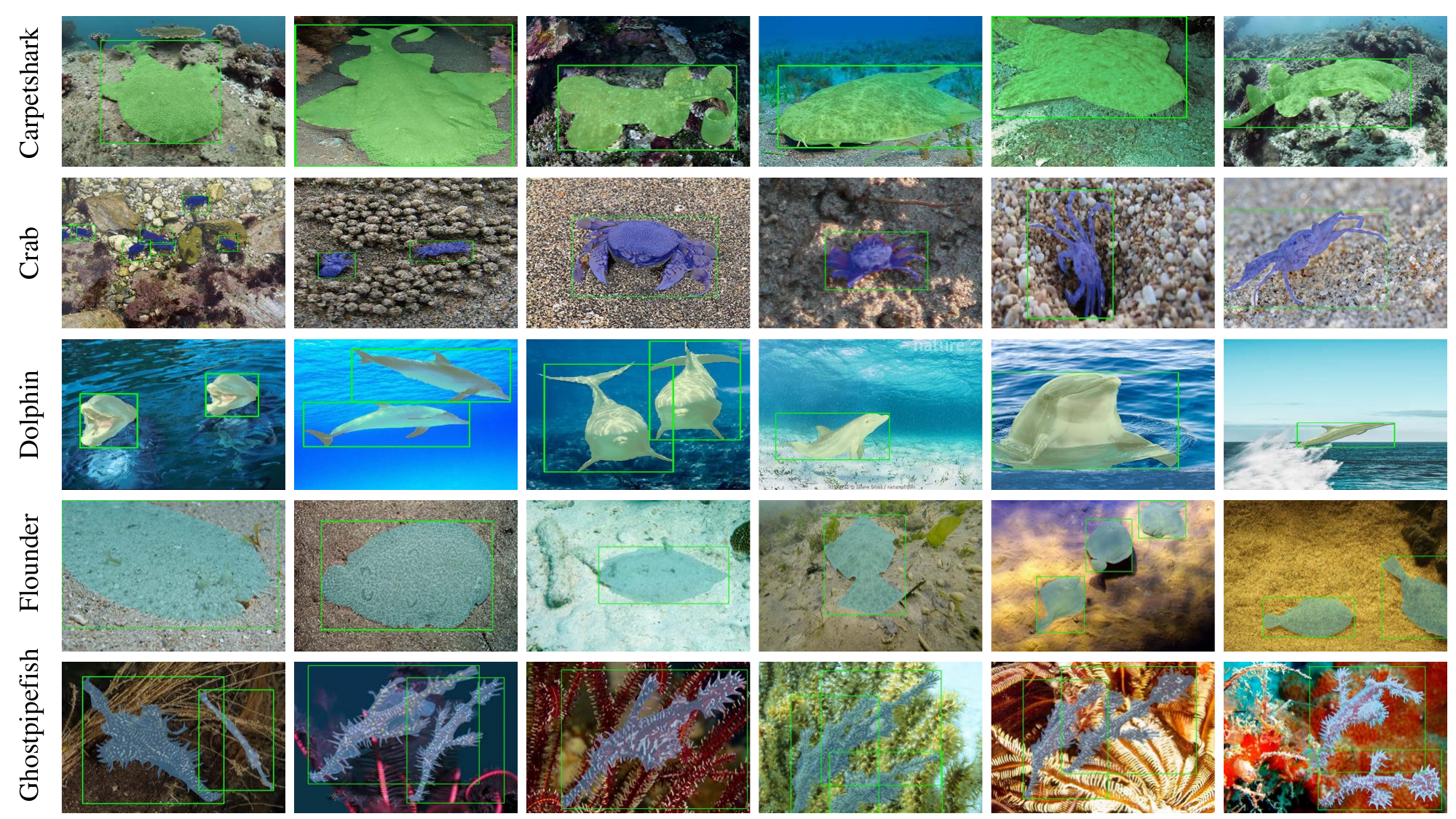}
    \includegraphics[width=\linewidth,page=2]{images2/dataset_samples_v1.pdf}
    \caption{Visualization of annotated images in our AAS dataset. The ground truth masks are overlaid onto the aquatic animal images.}
    \label{fig:dataset_samples}
\end{figure}

To the best of our knowledge, only three currently available animal datasets can be used for aquatic animal segmentation: subsets of the MAS3K~\cite{Li-Mas3K2020}, COD10K~\cite{Fan-CVPR2020}, and CAMO++~\cite{ltnghia-TIP2021} datasets. However, these datasets have different ranges of categories and a small number of images. This may lead to overfitting when training deep neural networks due to the small number of images in each category (see Table~\ref{table:datasets}).

Our goal is to build a comprehensive benchmark that can be used to evaluate deep learning models. Instead of creating an independent dataset by crawling images from the Internet and annotating them, we decided to spend more effort selecting images and labels from existing datasets so that our proposed dataset has something in common with others and can be utilized for different tasks, including conventional ones (the original purpose of existing datasets) and our newly created task (aquatic animal object detection and instance segmentation).

To reach this goal, we reorganized a new dataset by sampling the three datasets noted above and relabeling the sampled images. We first selected images containing aquatic animals from these datasets. Next, we manually discarded duplicate selected images and relabeled the images retained to match the categories in our dataset. Examples of annotated instances in our collected AAS dataset can be viewed in Figure~\ref{fig:dataset_samples}.


Our newly constructed Aquatic Animal Species (AAS) dataset contains $4,239$ images of aquatic animals in 46 categories. There are $5,041$ instances in our dataset, and each image has 1.2 instances on average and a maximum of 22 instances. Table \ref{table:datasets} shows the overall statistics for our AAS dataset compared with existing datasets. Figure \ref{fig:animal_category} illustrates animal categories in our AAS dataset. Fish takes the most significant ratio with $33$ species and $72\%$ number of images. Non-fish animals include $13$ species taking $28\%$ number of images. 

\begin{figure}[t!]
    \centering
    \begin{subfigure}[b]{0.525\textwidth}
        \centering
        \includegraphics[width=\linewidth]{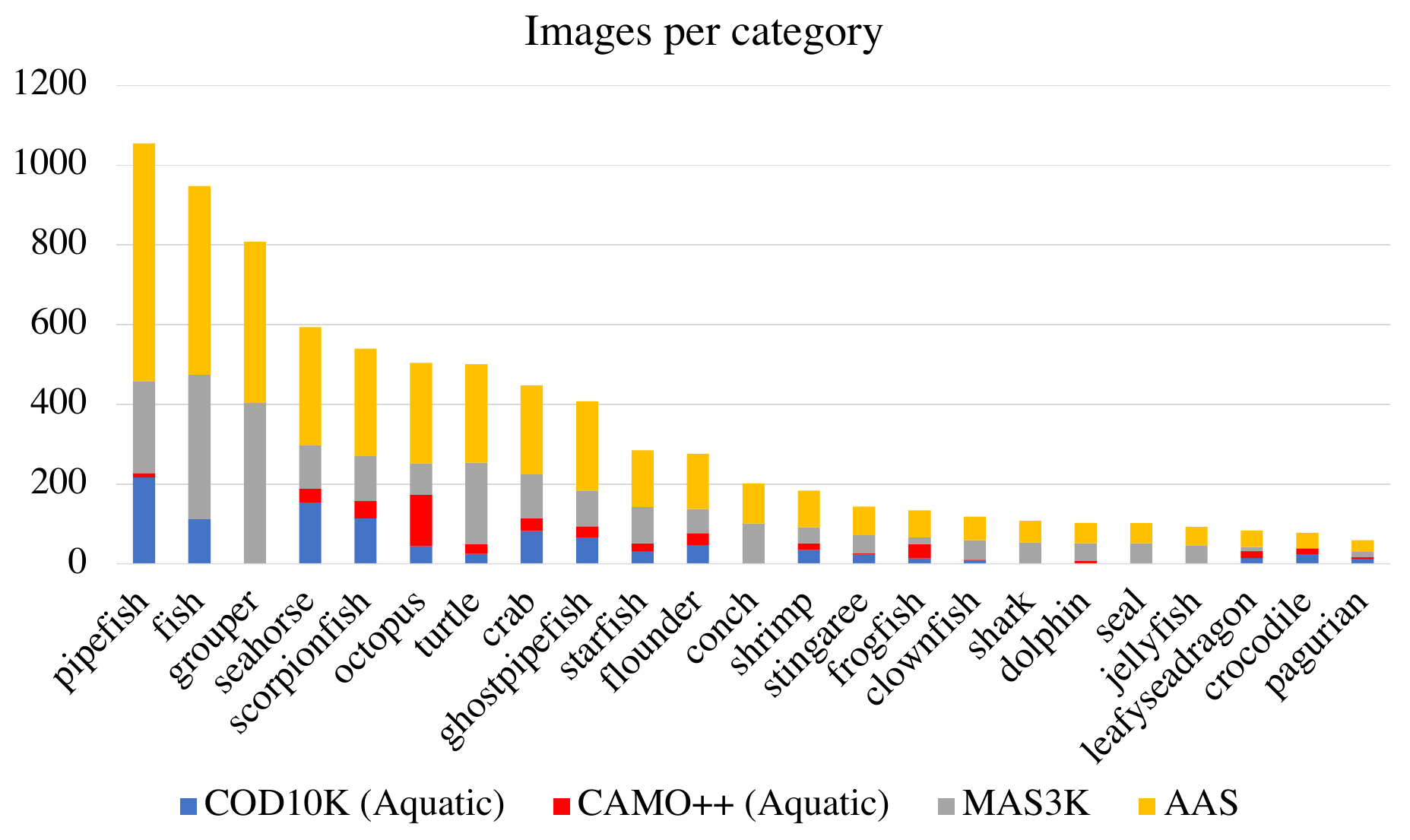}
    \end{subfigure}
    \hfill
    \begin{subfigure}[b]{0.46\textwidth}
        \centering
        \includegraphics[width=\linewidth]{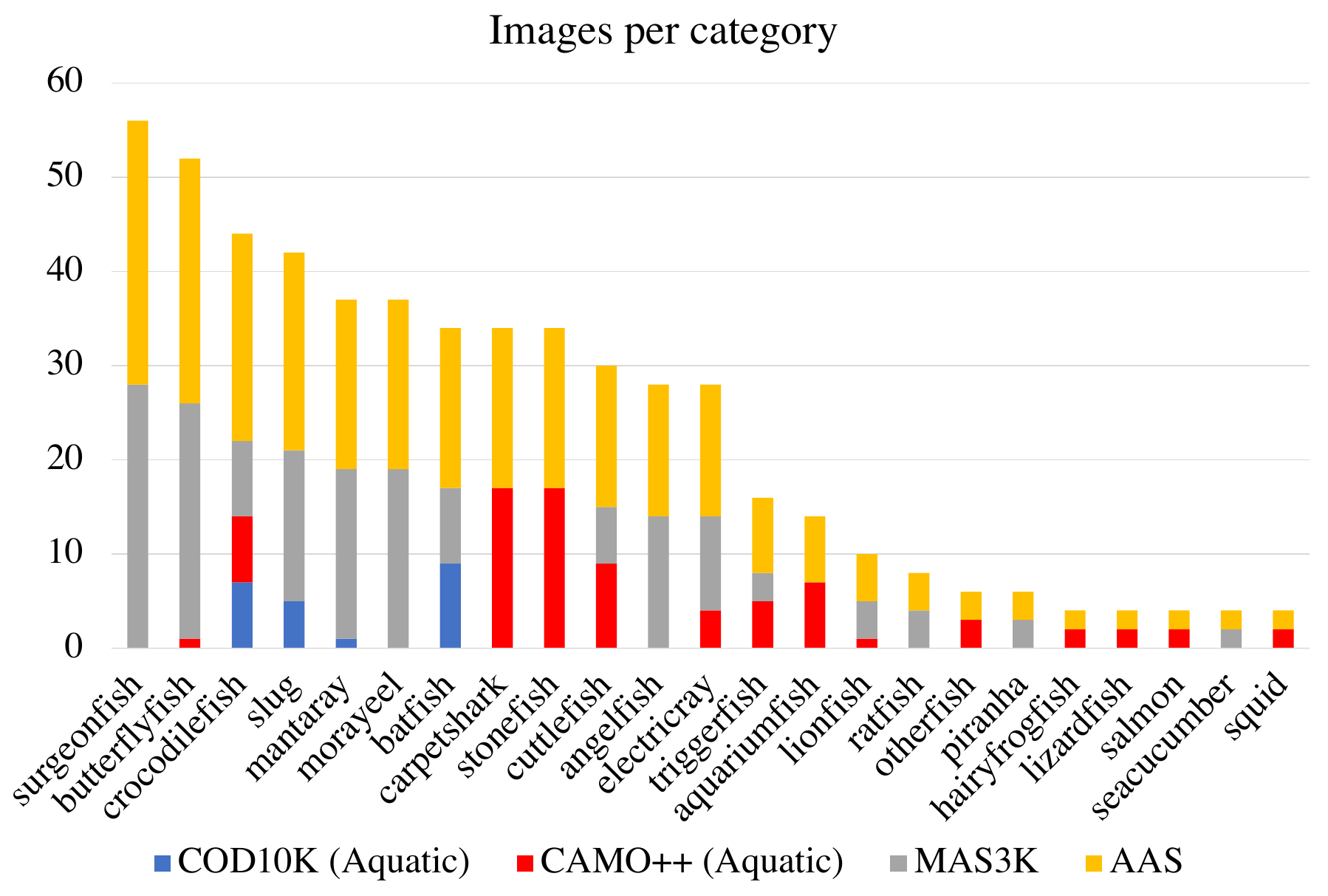}
    \end{subfigure}
    \caption{Number of images per category in our AAS, COD10K (Subset), CAMO++ (Subset) and MAS3K datasets.}
    \label{fig:images_per_cat}
\end{figure}

\begin{figure}[t!]
    \centering
    \begin{subfigure}[b]{0.48\textwidth}
    \includegraphics[width=\linewidth]{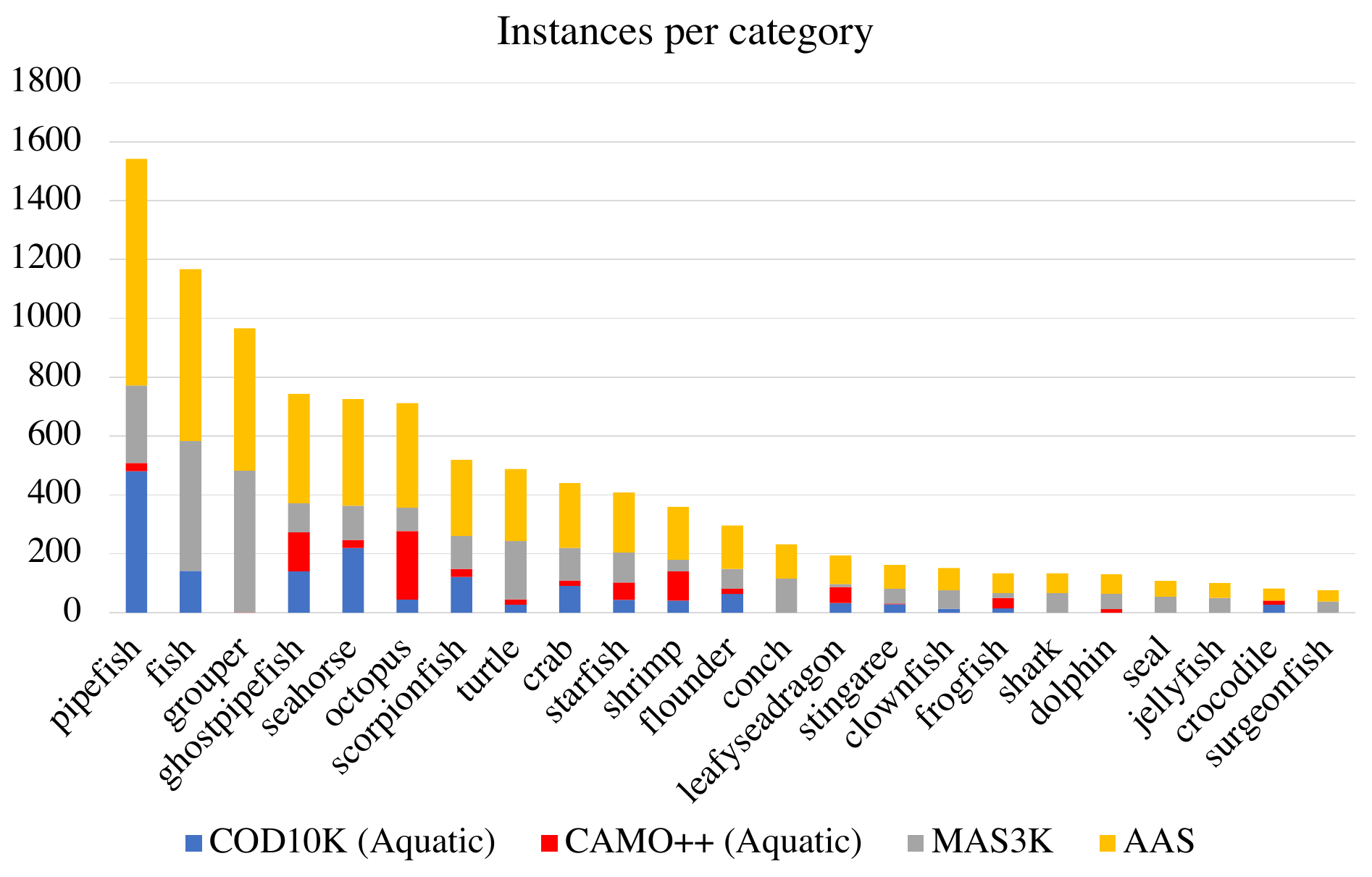}
    \end{subfigure}
    \begin{subfigure}[b]{0.495\textwidth}
        \centering
        \includegraphics[width=\linewidth]{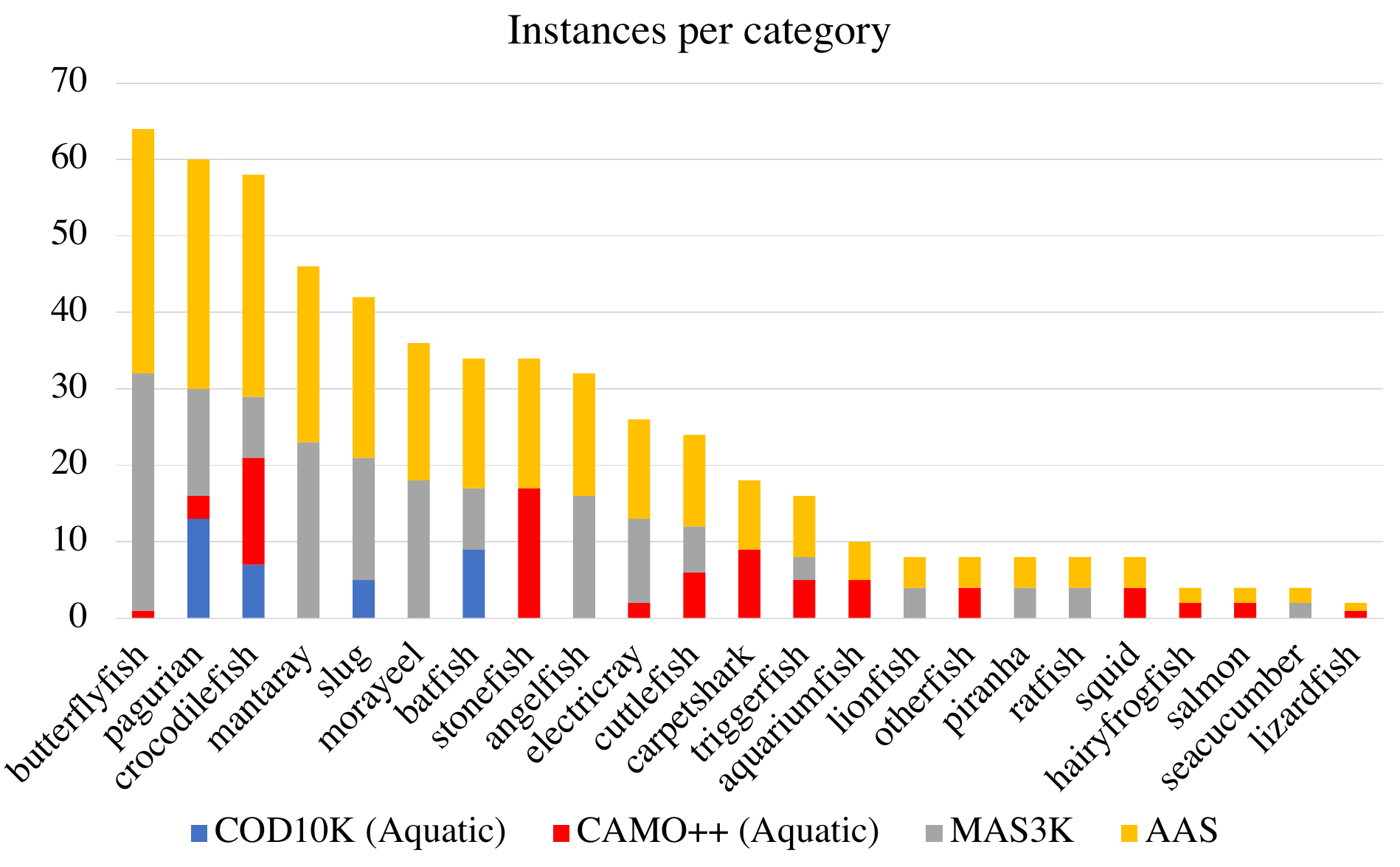}
    \end{subfigure}
    \caption{Number of annotated instances per category in our AAS, COD10K (Subset), CAMO++ (Subset) and MAS3K dataset.}
    \label{fig:instances_per_cat}
\end{figure}

The number of images and instances per category are illustrated in Figures~\ref{fig:images_per_cat} and \ref{fig:instances_per_cat}, respectively. Our collected AAS dataset has more aquatic categories than CAMO++, COD10K, and MAS3K datasets. Furthermore, concerning most categories, our proposed AAS has more images and instances than those prior datasets.

\textcolor{black}{We carefully selected and annotated images from existing datasets (MAS3K, COD10K, and CAMO++), adhering strictly to their respective licenses. Specifically, MAS3K images are publicly available for non-commercial research, COD10K images were released under the CC BY-NC-SA 4.0 license, and CAMO++ images are publicly available for academic and research purposes. We clearly attribute all sources and explicitly state modifications made. Our AAS dataset is accordingly released under a CC BY-NC-SA 4.0 license, compatible with original datasets, to encourage non-commercial academic use.}

To address overfitting caused by the small number of samples in certain categories, we implemented data augmentation techniques, such as guided mixup augmentation (Section 4.3) and standard regularization techniques (e.g., weight decay set to $10^{-4}$).

\section{Proposed Method}
\label{sec:proposed_method}

\begin{figure}[t!]
    \centering
    \includegraphics[width=0.85\linewidth]{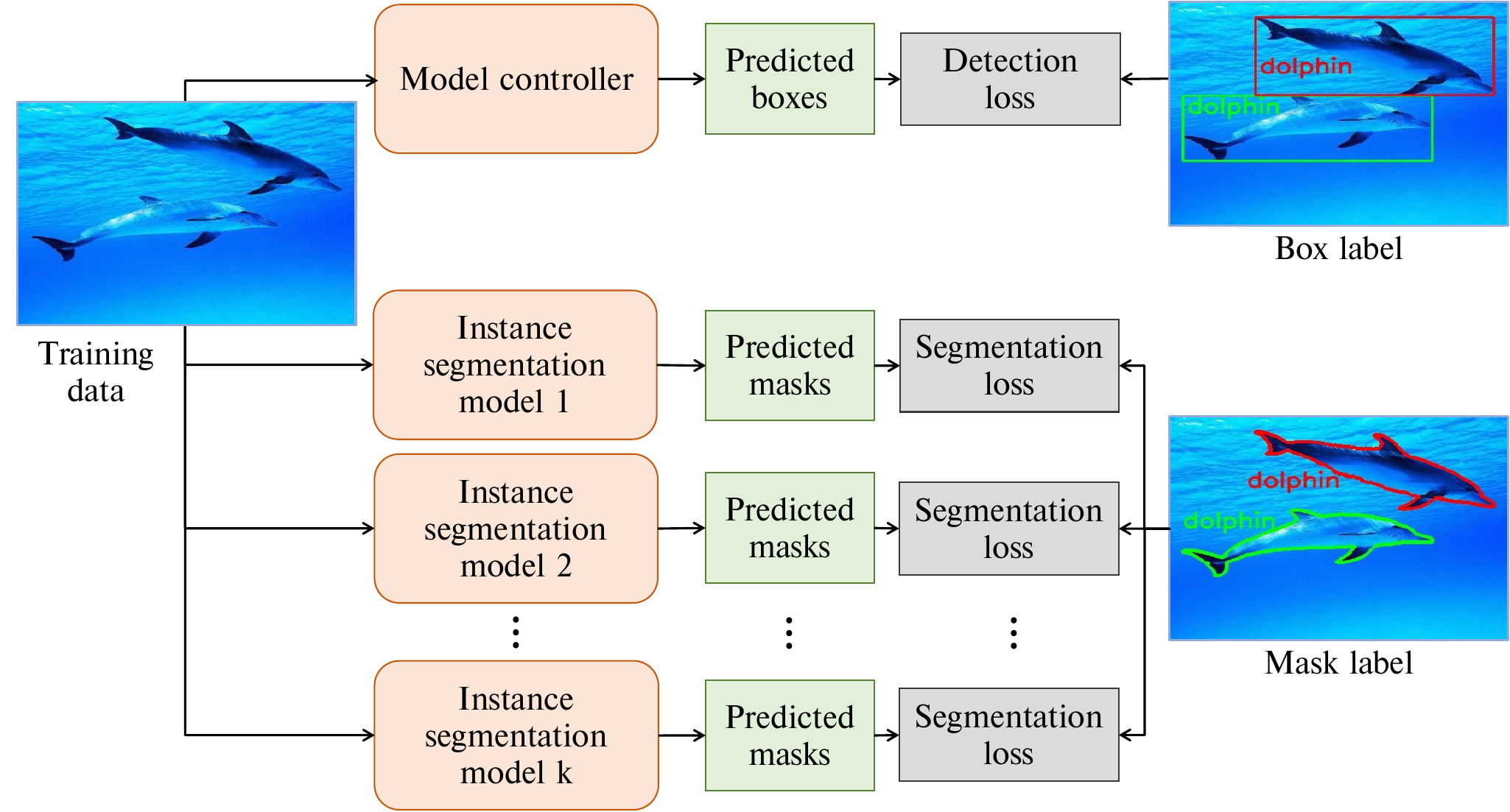}
    \caption{The training stage of our multi-model fusion method (best viewed online in color with zoom).}
    \label{fig:training}
\end{figure}

\subsection{Overview}

Instance segmentation methods~\cite{Kaiming-ICCV2017, Huang-CVPR2019} are imperfect because each method may have advantages in specific contexts but disadvantages in others. To utilize the strength of different methods, we derived a simple yet efficient \textbf{GU}ided mixup augme\textbf{N}tatio\textbf{N} and multi-mod\textbf{E}l fusion for aquatic anima\textbf{L} segmentation (GUNNEL) to instruct the training procedure and leverage various models through mining image contexts.

Figure~\ref{fig:training} illustrates the training stage of our proposed GUNNEL framework. In this stage, instance segmentation models and a detection model dubbed the ``model controller" are trained separately. The object detection-based model controller is trained with two loss components, including classification and localization losses.
\begin{equation}
    \mathcal{L} = \mathcal{L}_{\mathrm{cls}} + \mathcal{L}_{\mathrm{box}}
\end{equation}

Likewise, single-instance segmentation is trained with a multi-task loss function combining the loss of classification, localization, and segmentation.
\begin{equation}
    \mathcal{L} = \mathcal{L}_{\mathrm{cls}} + \mathcal{L}_{\mathrm{box}} + \mathcal{L}_{\mathrm{mask}}
\end{equation}

The single models exploit different contexts to create different segmentation masks for the aquatic animals in the image. The model controller aims to produce pseudo bounding box ground truths for fusing multiple instance segmentation models (see Section~\ref{sec:fusion}). 

\begin{figure}[t!]
    \centering
    \includegraphics[width=0.85\linewidth]{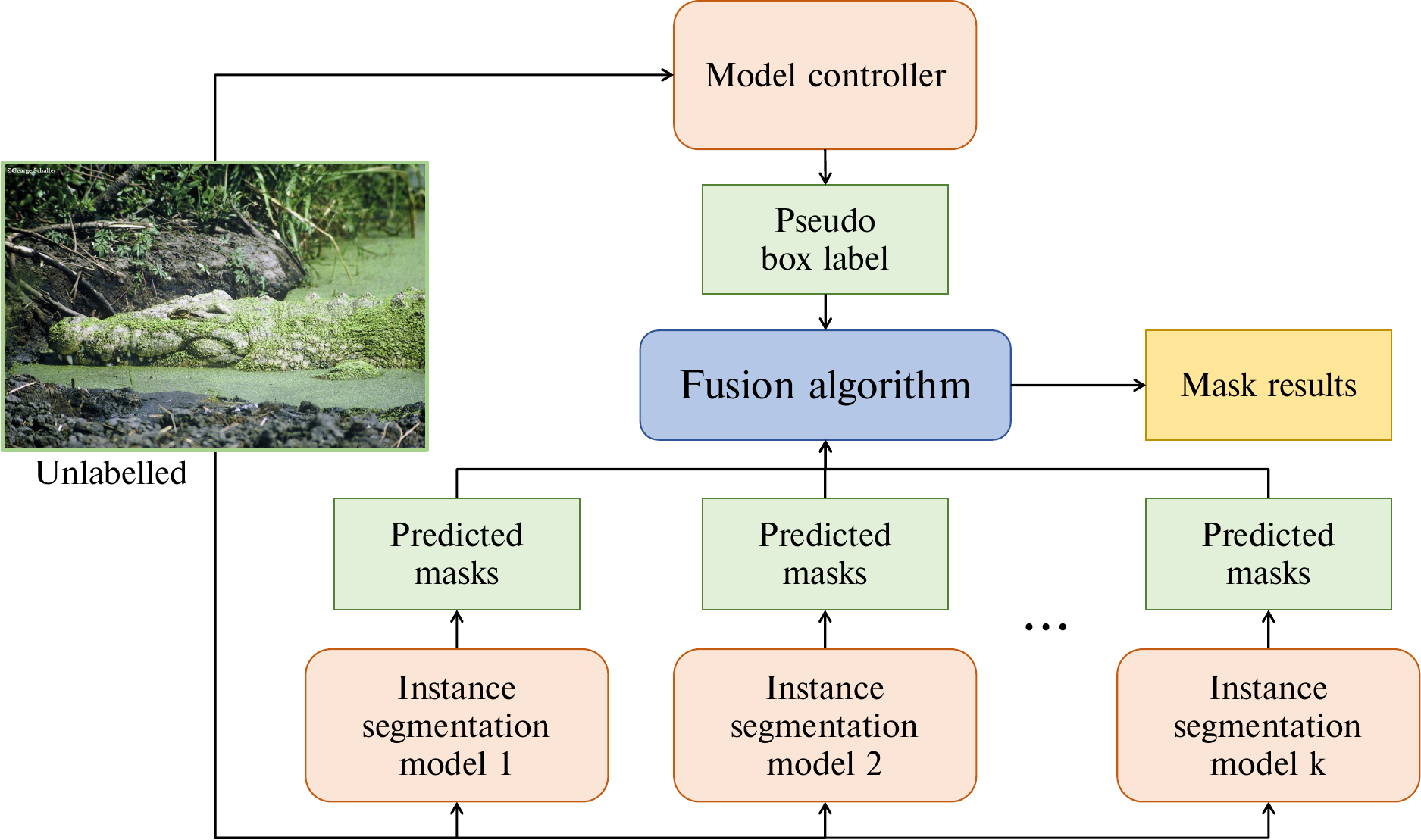}
    \caption{The inference stage of our multi-model fusion method (best viewed online in color with zoom).}
    \label{fig:inference}
\end{figure}

As shown in Figure~\ref{fig:inference}, in the inference stage, given an unlabeled image, the model controller extracts bounding boxes with high confidence scores from each image. These boxes guide the model controller to fuse the results from single models,  resulting in the final segmentation mask results.

\begin{algorithm}[t!]
\caption{Multi-model fusion algorithm.}\label{alg:TS_mask_fusion}
\begin{algorithmic}[1]
\Require{{$ $}
\begin{enumerate}
    \item Collection of instance segmentation results $\mathrm{ins}[N \times K]$ for $K$ single models on $N$ images.
    \item List of pseudo ground truths $\mathrm{ pseudo_{box}}[N]$ for $N$ images from model controller.
\end{enumerate}
}
\Ensure{{$ $}
    List of results $\mathrm{res}[N]$ for $N$ images. Each element of $\mathrm{ins}$, $\mathrm{res}$ contains all segmented results of the corresponding image.
}
\State $\textrm{res} \gets \emptyset$ 
\For{$i \gets 1$ to $N$}
\State {$\hat{k} \gets \arg\!\max\limits_{k \in K}(\mathrm{\textbf{AP}_{box}}(\mathrm{pseudo_{box}}[1:i], \textrm{res} \cup \textrm{ins}[i, k]))$}
\State {$ \textrm{res} \gets \textrm{res} \cup \textrm{ins}[i, \hat{k}]$}
\EndFor
\State \textbf{return} $\textrm{res}$
\end{algorithmic}
\end{algorithm}



\subsection{Multi-Model Fusion}
\label{sec:fusion}

Our proposed multi-model mask fusion algorithm uses a mask-independent detector as the guidance model controller. The YOLOv5~\cite{GlennJocher-Zenodo2020} model is used as the model controller thanks to its fast training and focusing on both spatial and discriminative representation.

\textbf{Instance Segmentation Models.} We used 10 widely known architectures: Mask RCNN~\cite{Kaiming-ICCV2017}, Cascade Mask RCNN~\cite{Zhaowei-CVPR2018}, GN-Mask RCNN~\cite{Wu-ECCV2018}, Mask Scoring RCNN~\cite{Huang-CVPR2019}, Deformable ConvNets~\cite{Xizhou-CVPR2019}, GCNet~\cite{Cao-CVPRW2019}, CARAFE~\cite{Jiaqi-ICCV2019}, WS-Mask RCNN~\cite{Siyuan-2019}, PointRend~\cite{Kirillov-CVPR2020}, and GRoIE~\cite{Rossi-ICPR2021}. Each model used the ResNet50 backbone integrated with a Feature Pyramid Network (FPN). All models were fine-tuned on the AAS dataset, initialized with MS-COCO pre-trained weights~\cite{Lin-ECCV2014}.

\textbf{Model Controller (YOLOv5).} The YOLOv5~\cite{GlennJocher-Zenodo2020} model controller consisted of CSPDarknet53~\cite{wang2020cspnet} as the backbone, PANet~\cite{liu2018path} for multi-scale feature fusion, and YOLO layers for bounding box predictions. Training parameters included a batch size of 16, a learning rate of 0.01 (cosine annealing schedule), and weight decay of $5 \times 10^{-4}$.
\textcolor{black}{We selected YOLOv5 due to its excellent balance between speed and accuracy, effective spatial feature extraction, and robustness in object detection tasks. While newer transformer-based models offer strong global contextual modeling and might potentially enhance performance, their suitability as fusion controllers remains an interesting direction for future research.}

Algorithm \ref{alg:TS_mask_fusion} illustrates our proposed multi-model fusion algorithm, which receives pseudo bounding box ground truths and segmentation masks from single models as input and generates final mask results. It iterates over the images to update a ``waiting list." At each iteration, the segmentation masks of the single models for an image are added to the waiting list, and an evaluation is performed to choose the best results.  

In particular, when the algorithm works with the $i^{th}$ image, the ``result list" contains the segmentation masks for the first $i-1$ images. The waiting list is the union of the result list and a temporary segmentation mask for the $i^{th}$ image. The bounding box Average Precision ($\mathrm{AP}_{box}$) between the pseudo ground truths and the waiting list is then evaluated. If the $n^{th}$ segmentation model provides the highest AP value, its segmentation mask for the $i^{th}$ image is appended to the result list. 

\subsection{Guided Mixup Augmentation}

\begin{figure}[t!]
    \centering
    \includegraphics[width=\linewidth]{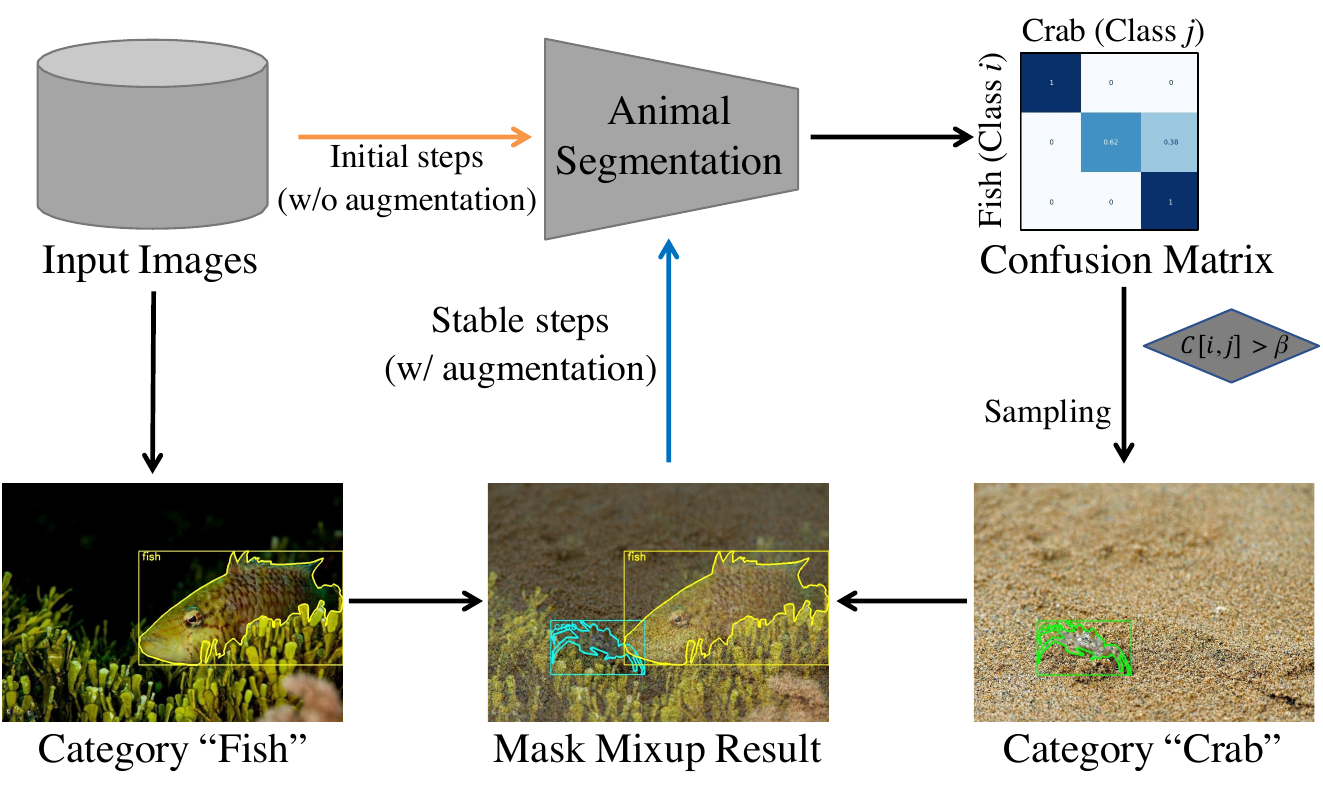}
    \caption{Workflow of proposed guided mixup augmentation using confusion matrix (best viewed online in color with zoom).}
    \label{fig:augmentation_framework}
\end{figure}

Aquatic animals involve both camouflaged and non-camouflaged ones, confusing deep learning models (see Section~\ref{sec:related_work_augmentation}). To address this problem, we developed a training procedure using guided mixup augmentation for instance segmentation. Unlike the mixup augmentation method~\cite{zhang-ICLR2018}, which chooses random pairs of images for blending, our wisely guided mixing strategy selects images using a confusion matrix. Figure~\ref{fig:augmentation_framework} shows the workflow of our proposed guided mixup augmentation, which involves two stages: warmup learning and augmentation learning.

The first few steps are for warmup learning: the single model (\ie, the instance segmentation model) tries to learn and adapt to new data by learning from only original images. Therefore, data augmentation is not used in these initial learning steps. 

\begin{figure}[t!]
    \centering
    \includegraphics[width=\linewidth]{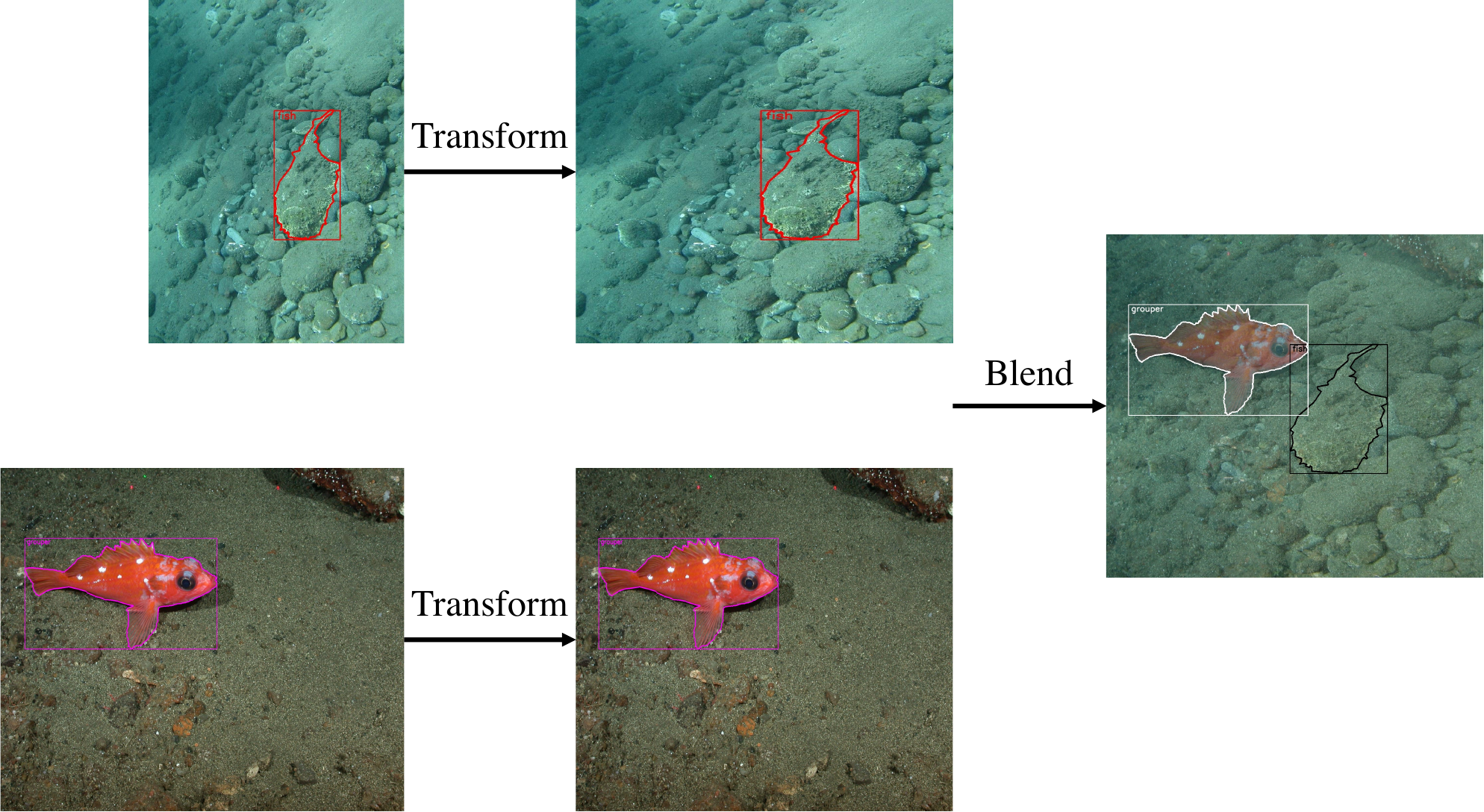}
    \caption{Blending process for data augmentation (best viewed online in color with zoom).}
    \label{fig:blending}
\end{figure}

Next, in the augmentation learning stage, when the single model achieves stable weights at the end of each epoch, a confusion matrix is constructed based on the bounding box IoU (intersection over union) between the segmentation results and ground truths. All instance ground truths with an IoU with the segmented instance mask larger than the threshold $\alpha=0.5$ are selected for each segmented instance instance. The category and classification score for all matching pairs of results and ground truths are extracted to create a guided confusion matrix $(C)$ for the next epoch.

In the next epoch, given an input image, we select an ambiguous image from the training set for mixing. In particular, for pairs of strongly misclassified categories $(i,j)$ (\ie, $C[i][j]$ is larger than the threshold $\beta=0.2$), if the input image contains instances of category $i$, a random image having instances in category $j$ is selected from the training set.

Figure \ref{fig:blending} illustrates our blending strategy. Given two images $a$ and $b$ with resolutions $(H_1, W_1)$ and $(H_2, W_2)$, we randomly resize the images within the range $(\frac{H_1+H_2}{2}\times[0.4, 0.6], \frac{W_1+W_2}{2}\times[0.4, 0.6])$ and then mix them together using $c = \gamma \times \phi (a) + (1-\gamma) \times \phi (b)$, where $\phi (\cdot)$ denotes random cropping and translation operations. Output image $c$ contains all instances in both original images, along with their ambiguous categories. Standard transformations, such as color jitter, brightness adjustment, and horizontal flipping, are applied to the output image. We empirically set $\gamma=0.5$. 

\textcolor{black}{Our guided mixup augmentation method specifically addresses challenging visual conditions such as poor lighting and blurry details by generating synthesized samples that are deliberately ambiguous, thus enhancing the robustness of the segmentation models. However, the performance of our method might still degrade in cases of extremely poor illumination or when encountering entirely unseen aquatic species. Such scenarios require further investigation and dedicated model adaptation techniques, which we identify as a future research direction.}

\subsection{Training Details}
The AAS dataset was divided into training (2,582 images), validation (516 images), and test sets (1,657 images). During training, $20\%$ of the training set was reserved for validation. The test set was strictly used for evaluation and was not part of training or validation. Instance segmentation models were trained using SGD optimization with momentum $0.9$ and weight decay $10^{-4}$. The learning rate was linearly warmed up from $2 \times 10^{-5}$ to $0.02$ and reduced by a factor of 10 at epochs 8 and 11.


\section{Experiments}
\label{sec:experiments}


\subsection{Implementation}

We used 10 instance segmentation models as the single models in our proposed GUNNE framework: Mask RCNN~\cite{Kaiming-ICCV2017}, Cascade Mask RCNN~\cite{Zhaowei-CVPR2018}, GN-Mask RCNN~\cite{Wu-ECCV2018}, Mask Scoring RCNN~\cite{Huang-CVPR2019}, Deformable ConvNets~\cite{Xizhou-CVPR2019}, GCNet~\cite{Cao-CVPRW2019}, CARAFE~\cite{Jiaqi-ICCV2019}, WS-Mask RCNN~\cite{Siyuan-2019}, PointRend~\cite{Kirillov-CVPR2020}, and GRoIE~\cite{Rossi-ICPR2021}. All of them were employed with the widely used ResNet50-FPN backbone. In addition, YOLOv5~\cite{GlennJocher-Zenodo2020} detector was used as the model controller.

All models were trained with a batch size of 2 on an RTX 2080Ti GPU. They were fine-tuned from their publicly released MS-COCO~\cite{Lin-ECCV2014} pre-trained models. 

The single segmentation models were trained using stochastic gradient descent (SGD) optimization~\cite{Bottou-COMPSTAT2010} with a weight decay of 
$10^{-4}$
and momentum of 0.9. 
The first 500 steps were used for warmup learning, in which the learning rate was set to 
$2 \times 10^{-5}$
and then linearly increased to 0.02. The models were then trained for 12 epochs with a base learning rate of 0.02, which was decreased by 10 times at the $8^{th}$ and $11^{th}$ epochs. Data augmentation was not applied in the first epoch so that the models could adapt to the aquatic animal images. From the second epoch, our proposed guided mixup augmentation method was applied. The augmented and original image ratio was kept at 50/50 using a Bernoulli distribution. 

The model controller was trained for $500$ epochs with an image size of $1280 \times 1280$, using default released parameters and our guided mixup data augmentation.


\subsection{Benchmarking}

In this section, we compare our GUNNEL framework with state-of-the-art methods to demonstrate the superiority of the proposed framework.

Table~\ref{table:result} shows the performance of different methods in terms of Average Precision (AP). As can be seen, our proposed GUNNEL achieved the best performance. Even without data augmentation, our method, which achieved an AP of $24.9\%$, still surpassed the state-of-the-art methods by a remarkable margin in all three AP metrics. Meanwhile, the completed GUNNEL achieved AP of $28.3\%$ and significantly outperformed all compared methods in all metrics. Furthermore, it shows the effectiveness of the proposed guided mixup augmentation, which results in an impressive increase of $13.7\%$. 

\begin{table}[!h]
\caption{Average Precision results of state-of-the-art instance segmentation methods and the proposed method; $1^{st}$ and $2^{nd}$ places are shown in \textcolor{blue}{\textbf{blue}} and \textcolor{red}{\textbf{red}}, respectively. MMF and GMA stand for Multi-model fusion and Guided mixup augmentation, respectively.}
\label{table:result}
\centering
\begin{tabular}{lccc}
\hline
\noalign{\smallskip}
\textbf{Method} & \textbf{AP} & \textbf{AP50} & \textbf{AP75} \\
\noalign{\smallskip}
\hline
\noalign{\smallskip}
Mask RCNN~\cite{Kaiming-ICCV2017} & 15.6 & 24.3 & 15.8 \\
Cascade Mask RCNN~\cite{Zhaowei-CVPR2018} & 20.6 & 30.2 & 22.2 \\
GN-Mask RCNN~\cite{Wu-ECCV2018} & 15.5 & 23.2 & 16.3 \\
Mask Scoring RCNN~\cite{Huang-CVPR2019} & 17.3 & 27.1 & 18.1 \\
Deformable ConvNets~\cite{Xizhou-CVPR2019} & 16.0 & 24.0 & 17.4 \\
GCNet~\cite{Cao-CVPRW2019} & 12.1 & 18.7 & 13.0 \\
CARAFE~\cite{Jiaqi-ICCV2019} & 14.7 & 22.7 & 15.2 \\
WS-Mask RCNN~\cite{Siyuan-2019} & 18.7 & 27.1 & 20.7 \\
PointRend~\cite{Kirillov-CVPR2020} & 12.7 & 18.7 & 13.3 \\
GRoIE~\cite{Rossi-ICPR2021} & 17.8 & 26.8 & 18.9 \\
DynaMask~\cite{li2023dynamask} & 22.4 & 31.8 & 24.6 \\
CamoFormer~\cite{yin2024camoformer} & 23.3 & 32.9 & 25.1 \\
\rowcolor{lightgray} Our GUNNEL (MMF) & \textcolor{red}{\textbf{24.9}} & \textcolor{red}{\textbf{36.3}} & \textcolor{red}{\textbf{26.5}} \\
\rowcolor{lightgray} Our GUNNEL (MMF + GMA) & \textcolor{blue}{\textbf{28.3}} & \textcolor{blue}{\textbf{40.7}} & \textcolor{blue}{\textbf{31.4}} \\
\hline
\end{tabular}
\end{table}

\begin{table}[!h]
\caption{Average Precision results of SOTA instance segmentation methods and the proposed method on CAMO++~\cite{ltnghia-TIP2021} dataset; $1^{st}$ and $2^{nd}$ places are shown in \textcolor{blue}{\textbf{blue}} and \textcolor{red}{\textbf{red}}, respectively.}
\label{table:result_camo++}
\centering
\begin{tabular}{lccc}
\hline
\noalign{\smallskip}
\textbf{Method} & \textbf{AP} & \textbf{AP50} & \textbf{AP75} \\
\noalign{\smallskip}
\hline
\noalign{\smallskip}
Mask RCNN~\cite{Kaiming-ICCV2017} & 16.1 & 36.4 & 12.9 \\
Cascade Mask RCNN~\cite{Zhaowei-CVPR2018} & 16.9 & 36.6 & 13.9 \\
GN-Mask RCNN~\cite{Wu-ECCV2018} & 16.2 & 36.1 & 13.2 \\
Mask Scoring RCNN~\cite{Huang-CVPR2019} & 19.1 & 38.1 & 18.0  \\
Deformable ConvNets~\cite{Xizhou-CVPR2019} & 16.5 & 36.2 & 13.3 \\
GCNet~\cite{Cao-CVPRW2019} & 12.8 & 32.6 & 12.2 \\
CARAFE~\cite{Jiaqi-ICCV2019} & 15.8 & 34.9 & 12.5 \\
WS-Mask RCNN~\cite{Siyuan-2019} & 20.1 & 38.4 & 19.7 \\
PointRend~\cite{Kirillov-CVPR2020} & 13.6 & 32.8 & 12.8 \\
GRoIE~\cite{Rossi-ICPR2021} & 19.3 & 38.0 & 18.4 \\
\rowcolor{lightgray} Our GUNNEL (MMF) & \textcolor{red}{\textbf{25.3}} & \textcolor{red}{\textbf{45.7}} & \textcolor{red}{\textbf{24.2}} \\
\rowcolor{lightgray} Our GUNNEL (MMF + GMA) & \textcolor{blue}{\textbf{29.6}} & \textcolor{blue}{\textbf{48.4}} & \textcolor{blue}{\textbf{30.6}} \\
\hline
\end{tabular}
\end{table}

\begin{figure}[!t]
\centering
\includegraphics[width=\textwidth, page=1]{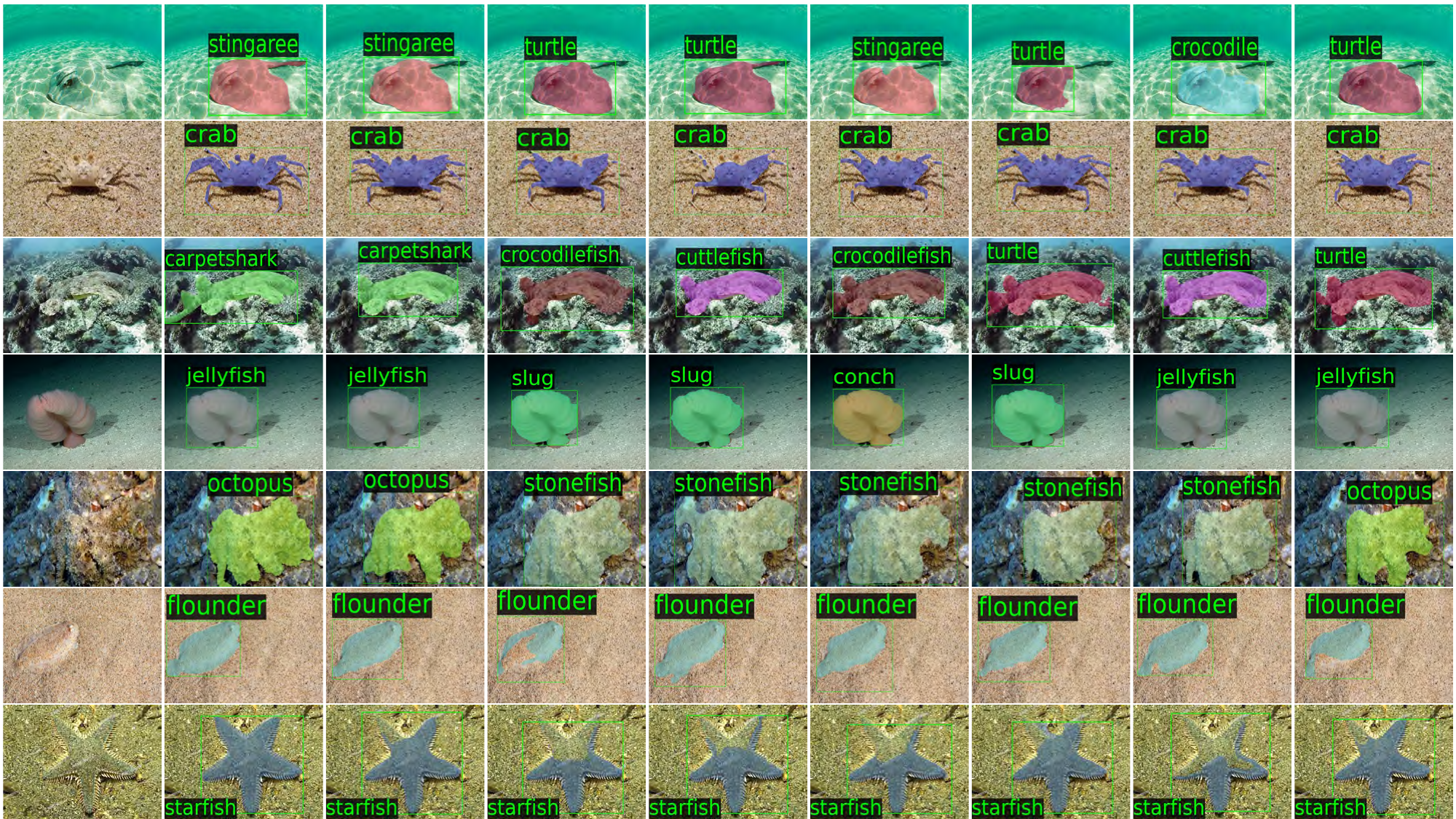}
\includegraphics[width=\textwidth, page=2]{images2/visualize_sub_v4.pdf}
\caption{Comparison of results of instance segmentation methods. From left to right: original image is followed by overlaid ground truth and results of our GUNNEL framework, Mask RCNN~\cite{Kaiming-ICCV2017}, Cascade Mask RCNN~\cite{Zhaowei-CVPR2018}, MS RCNN~\cite{Huang-CVPR2019}, PointRend~\cite{Kirillov-CVPR2020}, WS-Mask RCNN~\cite{Siyuan-2019}, GRoIE~\cite{Rossi-ICPR2021}. Instances shown in color differently from the ground truth denote misclassification. Best viewed in color with zoom.}
\label{fig:examples}
\end{figure}

Figure \ref{fig:examples} visually compares methods. Our GUNNEL achieved the best results close to the ground truth. Our method effectively handled both segmentation masks and classification labels.

\begin{table}[!t]
\caption{Effectiveness of data augmentation on the ecological version of GUNNEL. The best results are shown in \textcolor{blue}{\textbf{blue}}.}
\centering
\begin{tabular}{p{2in}p{0.5in}p{0.5in}p{0.5in}}
\hline
\noalign{\smallskip}
\textbf{Method} & \textbf{AP} & \textbf{AP50} & \textbf{AP75} \\
\noalign{\smallskip}
\hline
\noalign{\smallskip}
None & 21.7 & 32.5 & 22.5 \\
Copy-paste\cite{Ghiasi-CVPR2021} & 21.2 & 31.4 & 22.7 \\
Instaboost~\cite{Fang-ICCV2019} & 21.5 & 31.7 & 23.9 \\
Random mixup\cite{zhang-ICLR2018} & 22.1 & 32.0 & 24.2 \\
\rowcolor{lightgray} Our guided mixup & \textcolor{blue}{\textbf{23.8}}  & \textcolor{blue}{\textbf{34.5}} & \textcolor{blue}{\textbf{26.7}} \\
\hline
\end{tabular}
\label{table:augmentation}
\end{table}

We added CAMO++~\cite{ltnghia-TIP2021} to validate our approach. Table~
\ref{table:result_camo++} shows that GUNNEL outperforms all baseline methods across datasets, with notable improvements on camouflaged aquatic animals due to guided mixup augmentation and multi-model fusion.

\subsection{Comparison with Recent Methods}
We included comparisons with DynaMask~\cite{li2023dynamask} and CamoFormer~\cite{yin2024camoformer}. Results in Table~\ref{table:result} demonstrates that GUNNEL achieves state-of-the-art performance, particularly for camouflaged aquatic animals. DynaMask~\cite{li2023dynamask} demonstrates strong performance on non-camouflaged categories but struggled with camouflaged animals, as it lacked explicit mechanisms for addressing challenging backgrounds. CamoFormer~\cite{yin2024camoformer} excels in detecting camouflaged aquatic animals due to its advanced attention mechanisms, but it showed slightly lower generalization compared to our multi-model fusion approach in categories with high inter-class variability. Overall, GUNNEL achieves higher Average Precision (AP) scores across the board, particularly due to its guided mixup augmentation and robust fusion strategy.

\subsection{Ablation Study}

In this section, we analyze the core modules of the proposed GUNNEL framework to verify the effectiveness of each part. In particular, we took into account the effectiveness of guided mixup augmentation. We also investigated the number of single models and analyzed the complexity of our multi-model
fusion algorithm.

\subsubsection{Effectiveness of Guided Mixup Augmentation}

We demonstrated the usefulness of our proposed guided mixup augmentation method, with its category discrimination ability, by comparing its performance with that of other data augmentation methods, including random mixup~\cite{zhang-ICLR2018}, instaboost~\cite{Fang-ICCV2019}, and copy-paste~\cite{Ghiasi-CVPR2021}. We used public source codes of data augmentation methods released by the authors. We remark that due to limitations of existing data augmentation methods, we used an ecological version of our GUNNEL with two single models (\eg Mask RCNN~\cite{Kaiming-ICCV2017} and Cascade Mask RCNN~\cite{Zhaowei-CVPR2018}) to conduct the experiment efficiently.

As shown in Table \ref{table:augmentation}, our guided mixup augmentation substantially improved the performance of the GUNNEL framework by $9.7\%$. The performance of our proposed data augmentation method also surpassed that of existing data augmentation methods by a remarkable margin. Hence, our proposed guided mixup augmentation method can substantially contribute not only to reducing data labeling hours but also to efficiency (\ie, the processing time) and effectiveness (\ie, the performance improvement).

\subsubsection{Number of Single Models}

\begin{figure}[!t]
  \centering
  \centering
    \includegraphics[width=\linewidth]{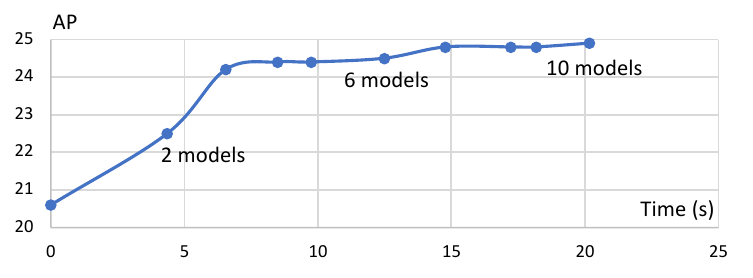}
  \caption{Trade-off between accuracy and processing time as the number of single models was increased. Accuracy plateaued at 10 models.}
  \label{fig:number_view}
\end{figure}
We investigated the effectiveness of our multi-model fusion by increasing the number of single models one by one. To show the actual contribution of the proposed fusion algorithm, we did not use our guided mixup data augmentation in the experiments. As shown in Fig.~\ref{fig:number_view}, the accuracy of the GUNNEL plateaued at $10$ models. Further increases in the number of single models would only lead to increased processing time without improving the performance. Hence, we use $10$ single models in our GUNNEL framework.

\subsubsection{Complexity of Multi-Model Fusion}

In this section, we analyze the space and time complexity of our multi-model fusion algorithm. As described in Section \ref{sec:fusion}, the multi-model fusion algorithm receives a pseudo bounding box from the model controller and predictions of $N$ images from $K$ single models. However, the excessive increase in the number of single models does not significantly improve fusion efficiency and effectiveness. Indeed, Figure \ref{fig:number_view} illustrates that the fusion results tend to saturate when the number of single models rises to 7, and it becomes plateauing at 10 models. On the other hand, inference time grows significantly. That is why we consider the number of single models constant and focus on analyzing the complexity of the fusion algorithm based on the number of images.

To begin with, we explore the time complexity of the multi-model fusion algorithm, which depends on the number of images. The main operator of the fusion algorithm is the AP evaluation inside the loop over $N$ images. The evaluation procedure of $i^{th}$ image also relies on $i-1$ previous ones. In particular, we calculate the number of evaluation operators, $n_{op}$, as the quantities of images increase: 
\begin{equation}
     n_{op} = 1 + 2 + \dots + n = \frac{n(n+1)}{2} \in \mathcal{O}(n^2),
\end{equation}
where $n$ is the number of total images in the test dataset. As a result, we conclude that the time complexity of the multi-model fusion algorithm is $\mathcal{O}(n^2)$.

Likewise, we further peruse the space complexity of the proposed multi-model fusion algorithm. In fact, the space resources expand linearly, corresponding to the number of images. The algorithm mainly takes up the space of $n$, multiplying the space of predictions on a single image, where $n$ is the number of total images. In short, the time complexity and the space complexity of our multi-model fusion algorithm are $\mathcal{O}(n^2)$ and $\mathcal{O}(n)$, respectively.

\section{Limitations and Future Work}
Despite the significant improvements demonstrated, our GUNNEL framework has certain limitations. First, the multi-model fusion inherently introduces additional computational complexity, which might limit real-time applicability. Second, performance under extremely adverse conditions such as severe image degradation, poor lighting, or completely unseen species requires further enhancement. Third, while our experiments focus on the AAS dataset and validate on CAMO++, future work should include broader cross-dataset evaluations to fully assess the generalization capability of GUNNEL. Exploring the integration of transformer-based methods and extending our framework to other domains or applications also present exciting directions for future research.

\section{Conclusion}
\label{sec:conclusion}
This paper has investigated the interesting yet challenging problem of aquatic animal segmentation. We created the Aquatic Animal Species (AAS) dataset containing images of diverse aquatic animal species. Each image was provided with annotated instance-level mask ground truths. We also developed a novel multi-model fusion algorithm to leverage different instance segmentation models. We further boosted the performance of aquatic animal segmentation by developing a guided mixup augmentation method. Extensive experiments demonstrated that our proposed method achieves state-of-the-art performance on our newly constructed dataset. We expect our AAS dataset to support significant research activities that automatically identify new aquatic animal species.

We plan to investigate various factors of the given problem in the future. Specifically, we will explore the use of contextual information in detecting and segmenting aquatic camouflaged instances. We will also investigate more different animals and plant life that live at varying sea depths. Furthermore, we intend to extend our work to dynamic scenes, such as underwater videos. We plan to investigate the impact of motion in segmenting aquatic camouflaged instances in videos. Additionally, exploring transformer-based architectures for further improving segmentation accuracy and robustness will be another exciting avenue of future work.




\section*{Conflict of Interest}

The authors declare that they have no known competing financial interests or personal relationships that could have appeared to influence the work reported in this paper.

\section*{Data Availability}

The dataset has been publicly available for academic purposes at \url{https://doi.org/10.5281/zenodo.8208877}. The code is available at \url{https://github.com/lmquan2000/mask-mixup}

\section*{Acknowledgements}

This work was funded by the Vingroup Innovation Foundation (VINIF.2019.DA19), the National Science Foundation Grant (NSF\#2025234), and the research funding from Faculty of Information Technology, University of Science, Vietnam National University - Ho Chi Minh City. Prof. Isao Echizen is supported by JSPS KAKENHI Grants JP21H04907 and JP24H00732, by JST CREST Grant JPMJCR20D3, by JST AIP Acceleration Grant JPMJCR24U3, and by JST K Program Grant JPMJKP24C2 Japan.



\balance
\bibliography{sn-bibliography}

\end{document}